\documentclass[11pt,a4paper]{article}
\usepackage[hyperref]{emnlp-ijcnlp-2019}
\usepackage{times}
\usepackage{latexsym}
\usepackage{import}
\usepackage{algorithm}
\usepackage[noend]{algpseudocode}
\usepackage{graphicx}
\usepackage{amsmath}
\usepackage{makecell}
\usepackage{stfloats}
\usepackage{color}
\usepackage{tabularx}

\usepackage{url}

\aclfinalcopy % Uncomment this line for the final submission

\title{Tackling Long-Tailed Relations and Uncommon Entities in Knowledge Graph Completion \thanks{The work described in this paper is substantially supported by a grant
from the Research Grant Council of the Hong Kong Special Administrative
Region, China (Project Code: 14204418).}}

\author{
Zihao Wang$^{1}$, Kwun Ping Lai$^1$, Piji Li$^{12}$, Lidong Bing$^3$, Wai Lam$^1$,\\
$^1$The Chinese University of Hong Kong\\
$^2$Tencent AI Lab, Shenzhen, China\\
$^3$R\&D Center Singapore, Machine Intelligence Technology, Alibaba DAMO Academy \\
zihaowangbupt@gmail.com \\
\{kplai, wlam\}@se.cuhk.edu.hk \\
pijili@tencent.com \\
l.bing@alibaba-inc.com
}

\date{}

\begin{document}

%\iffalse  
\maketitle
  
\begin{abstract}
	For large-scale knowledge graphs (KGs), recent research has been focusing on the large proportion of infrequent relations which have been ignored by previous studies.
For example few-shot learning paradigm for relations has been investigated.
In this work, we further advocate that handling uncommon entities is inevitable when dealing with infrequent relations.
Therefore, we propose a meta-learning framework that aims at handling infrequent relations with few-shot learning and uncommon entities by using textual descriptions.
We design a novel model to better extract key information from textual descriptions.
Besides, we also develop a novel generative model in our framework to enhance the performance by generating extra triplets during the training stage.
Experiments are conducted on two datasets from real-world KGs, and the results show that our framework \footnote{The implementation of our framework can be found in \url{https://github.com/ZihaoWang/Few-shot-KGC}.} outperforms previous methods when dealing with infrequent relations and their accompanying uncommon entities.
\end{abstract}

\section{Introduction}
\begin{figure}
	\centering
	\includegraphics[width=1.0\linewidth]{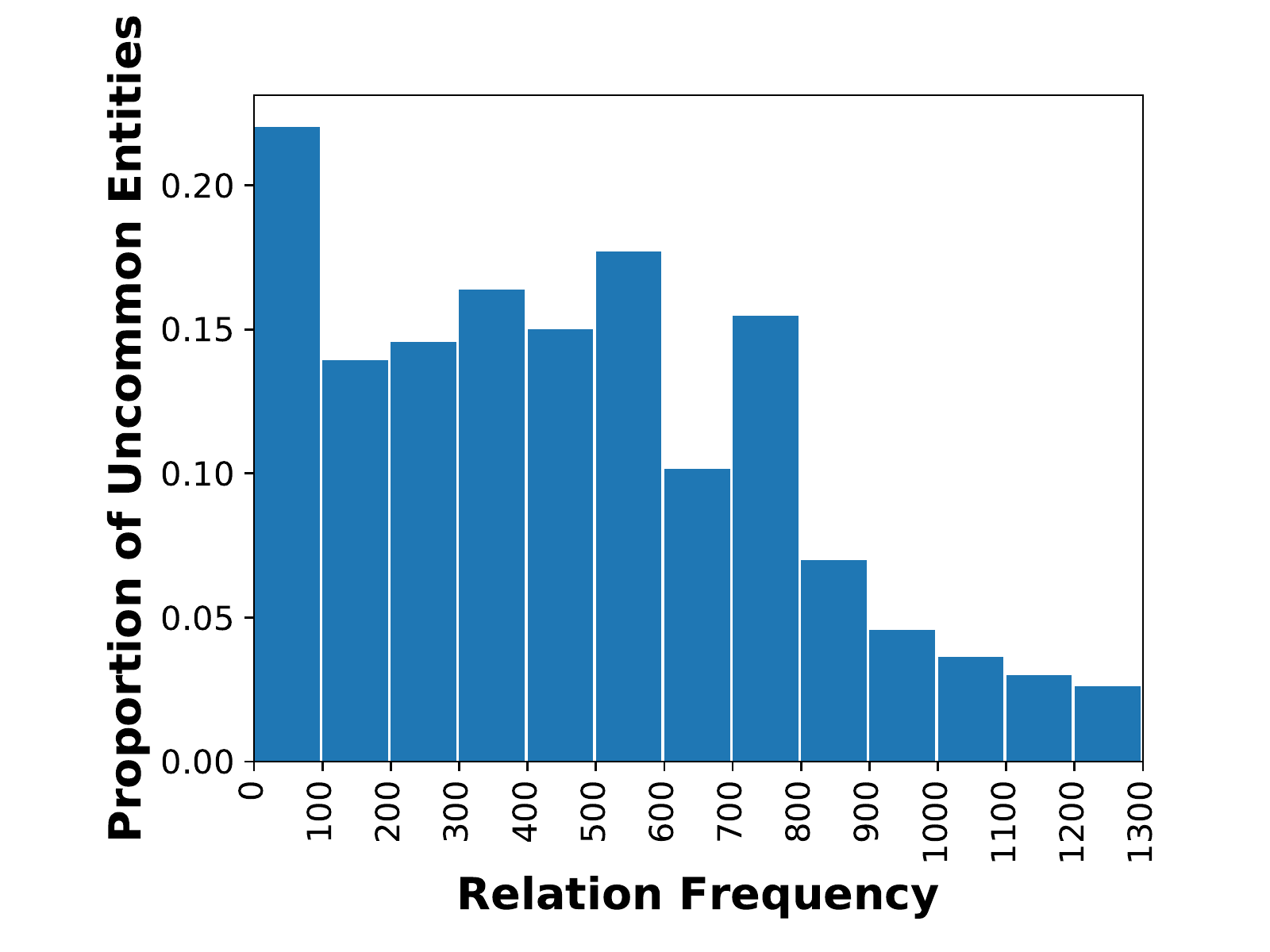}
	\caption{A histogram about relation frequencies and the corresponding proportions of uncommon entities in DBpedia.}
	\label{fig:100-14-6}
\end{figure}

Modern knowledge graphs (KGs)\cite{Bollacker:2008:FCC:1376616.1376746, lehmann2015dbpedia, Vrandecic:2014:WFC:2661061.2629489} consist of a large number of facts, where each fact is represented as a triplet consisting of two entities and a binary relation between them.
KGs provide rich information and it has been widely adopted in different tasks, such as question answering \cite{YihCHG15},  information extraction \cite{BingDMPC17,BingCWC15,BingLWC16} and image classification \cite{8099493}.
However, KGs still have the issue of incomplete facts.
To deal with the problem, Knowledge Graph Completion (KGC) task is introduced to automatically deduce and fill the missing facts.
There exist many previous works focusing on this task and embedding-based methods \cite{Bordes:2013:TEM:2999792.2999923,Wang:2014:KGE:2893873.2894046,Trouillon:2016:CES:3045390.3045609} achieves the best performance among them.
Recent works such as \cite{XiongYCGW18} have pointed out that relations in KGs follow a long-tailed distribution.
To be more precise, a large proportion of relations have only a few facts in KGs.
However, previous works of KGC usually focused on small proportions of frequent relations and ignored the remaining ones.
One observation is that they often conducted experiments on small datasets such as FB15k and WN18 \cite{Bordes:2013:TEM:2999792.2999923} where a relation typically possesses thousands of facts.
Moreover, after analyzing real-world KGs, we find that the more infrequently a relation appears, the entities within its facts are also more uncommon.
Figure \ref{fig:100-14-6} depicts the relationship between the relation frequency and the proportion of uncommon entities that appear in the facts of these relations in a KG, where an entity is treated as uncommon when it appears less or equal than 5 times in all triplets of the KG.
From Figure \ref{fig:100-14-6}, it is obvious that less frequent relations involve more uncommon entities than frequent relations. 
Therefore, when dealing with the problem of infrequent relations, the issue of uncommon entities should also be considered simultaneously, where they are two sides of a coin.

Previous works such as \cite{XiongYCGW18} only focused on those infrequent relations and ignored the accompanying problem of uncommon entities.
When handling uncommon entities, relying only on the structural information of KGs would lead to inferior performance due to data insufficiency, and thus additional information is required.
Some works \cite{toutanova-etal-2015-representing,XieLJLS16} utilize textual description of entities, but they cannot extract different information from entity description if the entity is involved in more than one relations.
A recent work \cite{ShiW18} tries to tackle this problem by using an attention mechanism considering both entity description and relation, but it adopts a heuristic method that cannot generalize well.

In this paper, we consider performing KGC for infrequent relations and uncommon entities as a few-shot learning problem, and we propose a framework that consists of three main components: description encoder, triplet generator, and meta-learner.
In the description encoder, we design a novel structure to handle entities involved with multiple relations by automatically locating and extracting relation-specific information.
We also simultaneously learn a triplet generator that is able to generate extra triplets in order to relieve the problem of data sparsity in few-shot learning.
Moreover, a meta-learner is further adopted to learn a initial representation of the model that can be easily adapted to unseen relations and entities.
As a result, our work has three main contributions as follows:
\begin{itemize}
	\item We formulate the problem of infrequent relations and uncommon entities as a few-shot learning problem and propose a meta-learning framework to solve it.
	
	\item We propose a novel model to extract relation-specific information from entity description for entities with multiple relations.
		
	\item We propose a generative model that can enhance the performance of few-shot learning by generating extra triplets during the training stage.  	
\end{itemize}

\section{Related Work}
\subsection{Knowledge Graph Completion}
The knowledge graph completion (KGC) task focuses on automatically inferring and filling the missing facts in KG.
The most successful method of KGC is the embedding-based method that learns a latent embedding in a common space for entities and relations.
This method only relies on the structural information of entities and relations in KG.
There exist a variety of methods \cite{Bordes:2013:TEM:2999792.2999923, Socher:2013:RNT:2999611.2999715, Wang:2014:KGE:2893873.2894046, Trouillon:2016:CES:3045390.3045609, nguyen-etal-2018-novel} that have been proposed to learn good embeddings for entities and relations.

However, embedding of uncommon relation or entities can not learn a good representation due to the data insufficiency.
Some research has proposed that additional information can be introduced to enhance the learning performance.
Among different types of information, textual descriptions is commonly considered by previous works \cite{Zhong2015AligningKA, toutanova-etal-2015-representing, XieLJLS16, ShiW18}.
Recently, meta-learning is also proposed by \cite{XiongYCGW18} to learn infrequent long-tailed relations in KG.

\subsection{Meta-Learning}
Meta-learning \cite{Lemke2013MetalearningAS} aims at learning common experiences across different tasks and easy adapting the existing model to new tasks.
One interesting application of meta-learning is few-shot learning problem where each task has only a few training data available.

Some research focus on learning a general policy for different tasks using a neural network.
An early work \cite{Santoro:2016:MMN:3045390.3045585} proposes that the learning policy can be learned by using a global memory network.
Recently, temporal convolution and attention have been considered to learn a common representation and pinpoint common experiences \cite{Mishra2018ASN}.

Another direction is to learn a good initial representation where the learned model can be easily adapted to new data.
Prototypical Network \cite{Snell2017PrototypicalNF} is proposed to learn a prototype for each category, and thus new data can be classified by distances between data and prototypes.
Model-Agnostic Meta-Learning (MAML) \cite{FinnAL17} focuses on learning a good initial point in parameter space of model, hence a trained model can be quickly adapted to new tasks with several updates.
More recently, Reptile \cite{nichol2018first} proposes to be an approximation of MAML.
In the Reptile model parameters are updated after a number of steps of inner iteration that can maximize within-task generalization.

\section{Background and Overview}
\subsection{Problem Setting}
Knowledge graph (KG) consists of a set of facts.
Each fact has the form of a triplet $(h, r, t)$ where $h$ is a head entity, $r$ is a relation and $t$ is a tail entity.
KGs are usually sparse, incomplete, and noisy.
Therefore Knowledge Graph Completion (KGC) becomes an important task.
Given arbitrary two out of three elements within a triplet, the goal of KGC is to predict the remaining one.
We focus on predicting $t$ given $h$ and $r$ in this work since our purpose is to deal with uncommon entities.

Previous works usually considered performing KGC given a set of common relations with lots of triplets.
On the contrary, we concentrate on performing KGC on those relations that have only a small number of triplets, which can be viewed as a $k$-shot learning problem for relations when $k$ is a small number.
In the limiting case where $k$ equals to 1, we deal with one-shot learning problem in our framework.
Besides, unlike the previous work \cite{XiongYCGW18} that focuses on common entities, we also consider uncommon entities during operational or testing phase, which means that some entities could appear only several times or be absent before.
Moreover, when dealing with uncommon entities, relying only on the structural information of KG would lead to inferior performance due to data insufficiency, and thus additional information is necessary.
Textual descriptions have been widely considered in KGC.
Typically, textual descriptions are used to describe an entity or a relation, and each of them can be either a short sentence or a paragraph consisting of several sentences.
We utilize textual descriptions of entities and relations in our framework.

\subsection{Overview of Learning Method}
Meta learning is a popular paradigm for solving the few-shot learning problem, and we adopt it to perform KGC.
Given a KG, we treat each relation as a task, and the triplets of each relation can be viewed as specific data of each task.
We further divide all tasks into three disjoint sets $R_{train}$, $R_{val}$ and $R_{test}$. Hence meta-training, meta-validation and meta-testing can be performed on each set respectively.
In each iteration of the meta-training phase, we randomly sample $B$ tasks from $R_{train}$ where $B$ is batch size, and then for each task $r$ in the batch we sample some triplets of $r$ to train the model.
After meta-training finishes, we obtain a trained model with model parameters $W$.
Next, we follow the procedure in the previous work \cite{XiongYCGW18} to perform  meta-validation and meta-testing on $R_{val}$ and $R_{test}$, where the settings are the same.
So we only describe meta-testing for short.
In the meta-testing phase, given a new task $r^{'} \in R_{test}$ with $H_{r^{'}}$ triplets, we randomly sample $k$ out of $H_{r^{'}}$ triplets.
The trained model is further improved via another training stage with only these $k$ samples and the parameters of model become $W^{'}$.
Then we keep parameters being fixed as $W^{'}$ and evaluate the performance of model on the remaining $H_{r^{'}} - k$ triplets.
These procedures are repeated for all tasks in $R_{test}$.

Given a triplet, the textual descriptions of $h$, $r$ and $t$ are respectively $d_h$, $d_r$ and $d_t$.
With textual descriptions, entities and relations can be mapped into a common semantic space.
Therefore, uncommon entities can be tackled in this common space as usual.

\section{Model Description}
In this section we present the architecture and the learning procedure of our proposed framework.

First, given the textual descriptions of a triplet $(h, r, t)$, the description encoder extracts key information from descriptions and produces corresponding embeddings $O = (o_h, o_r, o_t)$, where $o$ is a $u$-dimensional vector in the latent semantic space.

Next, the triplet generator participates in the learning procedure.
During meta-training phase, it takes $O$ as inputs and learns latent patterns for triplets.
However, during the training stage of the meta-testing phase, instead of learning latent patterns, the triplet generator performs triplet augmentation by generating extra $K$ sets of embeddings $G = \{(g_{h}, g_{r}, g_{t})\}$.
Each set of embeddings $(g_{h}, g_{r}, g_{t})$ can be viewed as an artificial imitation of $O$.
In the few-shot setting, the size of $O$ is usually too small to learn a good representation for a new task, and thus extra embeddings $G$ is generated for data augmentation.

After previous procedures, we are able to obtain a set of embeddings of triplets $E = \{(e_h, e_r, e_t)\}$, where $E = O$ during meta-training phase and $E = O \cup G$ during training stage of meta-testing phase.
With $E$ prepared, a score function $F$ takes $E$ as input and computes the score $C$ for each group of embeddings $(e_h, e_r, e_t) \in E$.
Although more sophisticated score functions might be designed, in our framework we adopt a simple formula as follows:
\begin{equation}
C = F(h, r, t) = |e_h + e_r - e_t|
\label{eq:1}
\end{equation} from TransE \cite{Bordes:2013:TEM:2999792.2999923}, where $L_1$-norm is used.

Finally, during the meta-training phase or the training stage of the meta-testing phase, a loss function $L$ related to $C$ is computed, and the meta-learner is adopted to optimize $L$ so that the framework can be easily adapted to new relations and entities.
Otherwise, during the testing stage of the meta-testing phase, we collect scores of the correct triplet and other candidates, and then we compute metrics based on the rank of correct triplet within all scores for evaluation.

\subsection{Description Encoder}
\begin{figure}
	\centering
	\includegraphics[width=1.0\linewidth]{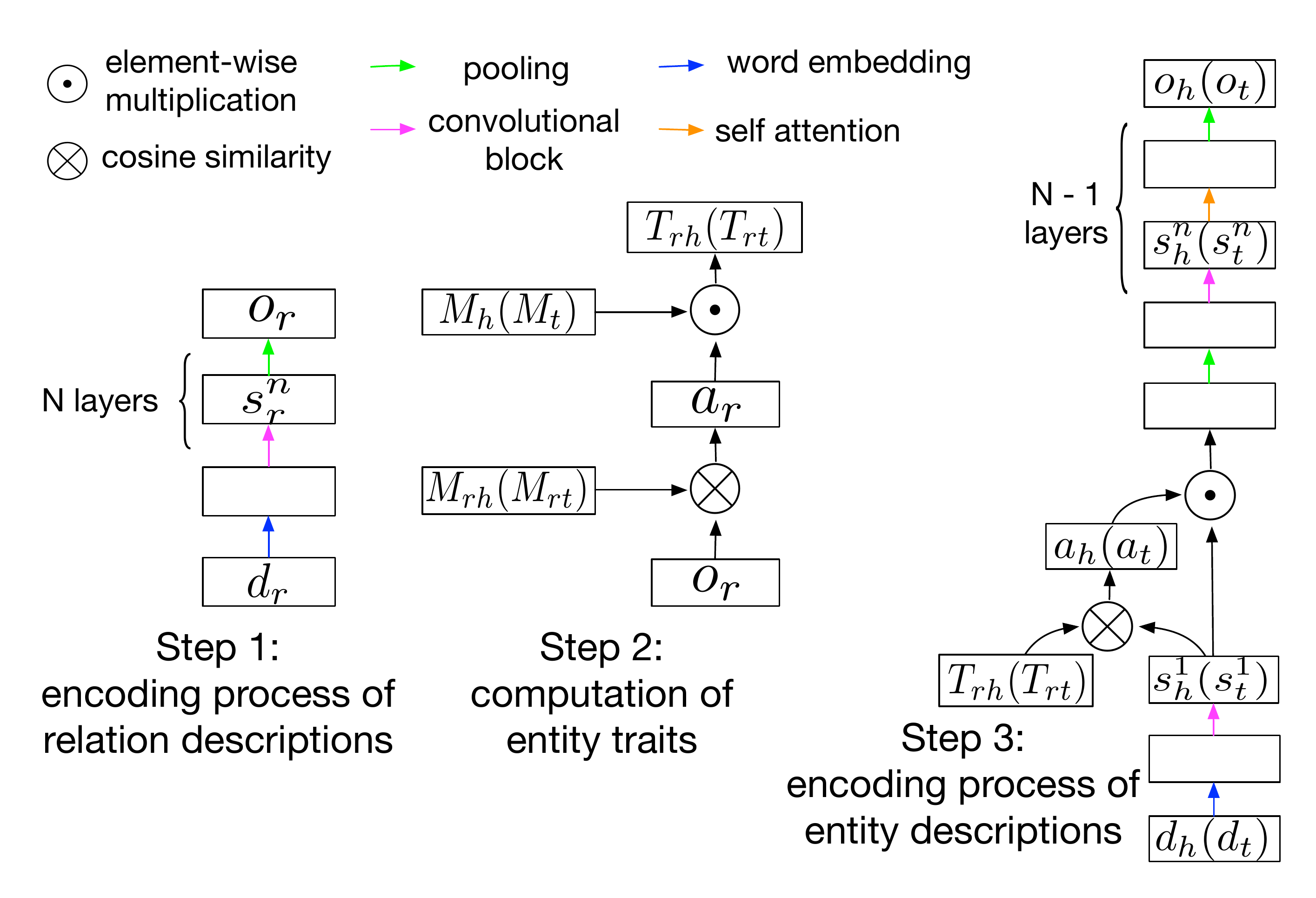}
	\caption{Structure and learning procedures of our description encoder. This figure should be viewed in color.}
	\label{fig:figure2}
\end{figure}

In KG, if an entity is involved in multiple relations, it is natural that different relations are more relevant to different parts in the description of the entity.
However, existing works using textual descriptions have not tackled this issue effectively.
In order to deal with this issue, we define a new concept "entity trait" ("trait" for short) that represents the common characteristics of some entities related to a special relation.
In another word, an entity owns different traits for different relations it involved.
In a sense, a trait is similar to an entity type ("type" for short), but it has more advantages when handling KGC.
First, types are not relation-specific but traits are.
Besides, a trait may consist of semantics of several different types and hence it is more expressive.
Moreover, we cannot easily obtain types in some situations, but traits can always be learned properly since they are latent and data-driven.
Formally, we assume that a relation $r$ has two traits $T_{rh}$ and $T_{rt}$, where the previous one for all the head entities of $r$ and the latter one for all the tail entities of $r$.
In our description encoder, a simple but effective method is adopted to learn and utilize traits to extract relation-specific information from description.

The overall structure and learning process of our description encoder are given in Figure \ref{fig:figure2}.
Given the descriptions $(d_h, d_r, d_t)$ of a triplet $(h, r, t)$, there are three steps to obtain the embeddings of triplet $O$ as depicted in the figure.
For entity, we only describe the process of $h$ for simplicity, but everything stays the same for $t$.
In Step 1, the encoding process of relation descriptions takes $d_r$ as input and outputs a relation embedding $o_r$.
Next, $o_r$ is used to learn the trait $T_{rh}$ for all the head entities of the relation $r$ in Step 2.
Finally, in Step 3 both $d_h$ and $T_{rh}$ are fed to the encoding process of the entity descriptions, and the output $o_h$ is the embedding of the head entity.
Note that the word embedding layer, convolutional blocks and pooling layers in Step 1 share the same parameters and architectures with the corresponding ones in Step 3.

The core of the encoding process is a $N$-layer convolutional neural network (CNN) \cite{conneau-etal-2017-deep}, which is shown to have excellent ability of extracting information.
In our CNN, the basic convolutional block consists of three consecutive operations: two 1-d convolutions, an instance normalization \cite{DBLP:journals/corr/UlyanovVL16}, and a non-linear mapping.
%We pad the input of all convolutions with proper number of zeros so that the output of 1-d convolution has the same shape as input.
For the pooling strategy, max pooling with a proper stride is used to distill the key information in the previous $N-1$ layers, and mean pooling is used to gather the information in the last layer.
Moreover, in Step 3, we also apply self-attention mechanism \cite{NIPS2017_7181} before each pooling layer in the last $N-1$ layers.
Unlike Step 1, self-attention is necessary here since entity descriptions are often more complex and noisier than relation descriptions according to our observation.
Self-attention can assign lower weights to noise, and then those noise would be filtered out in the subsequent pooling layer.

Furthermore, we demonstrate how to compute the trait in Step 2, where the external memories $M$ play an important role.
These memories record the global information of relations and entities that can generalize well when encountering new relations and entities.
In detail, the $o_r$ computed in Step 1 is transformed to a probability distribution $a_r$ by using $m$ relation memories $M_{rh}$ 
\begin{equation}
a_r = softmax(o_r \otimes M_{rh}),
\end{equation}
where $a_r$ is $m$-dimensional, $\otimes$ denotes the cosine similarity, $softmax$ is the commonly used softmax function and $M_{rh}$ is a matrix with shape $(m, u)$.
After that, the $u$-dimensional trait $T_{rh}$ can be obtained by computing a linear combination of $m$ latent entity memories $M_h$
\begin{equation}
T_{rh} = \sum_{i=1}^{m} M_h^i \odot a_r^i,
\end{equation}
where $\odot$ is the element-wise multiplication between two vectors and $M_h^i$ is a matrix with shape $(m, u)$.
Note that each pair of $m$ latent relation memories $M_{rh}$ and $m$ latent entity memories $M_h$ has a one-to-one correspondence.

Finally, we describe how the trait $T_{rh}$ is used to extract the relation-specific information in Step 3.
Given the description $d_h$, the hidden states $s^1_h$ can be obtained after the first convolutional block, and then the trait $T_{rh}$ is used to locate important hidden states in $s^1_h$ that have high relevance to $r$ by assigning them higher weights.
The procedure here is the same as before: a probability distribution $a_h$ over $s^1_h$ is computed by
\begin{equation}
a_h = softmax(T_{rh} \otimes s^1_h),
\end{equation}
and then $a_h$ multiplies with $s^1_h$ element-wise to weight different hidden states.
In this way, the hidden states that are not relevant to $r$ are assigned lower weights, and thus they are more likely to be filtered in the subsequent max-pooling layer.

\subsection{Triplet Generator}
\begin{figure}
	\centering
	\includegraphics[width=1.0\linewidth]{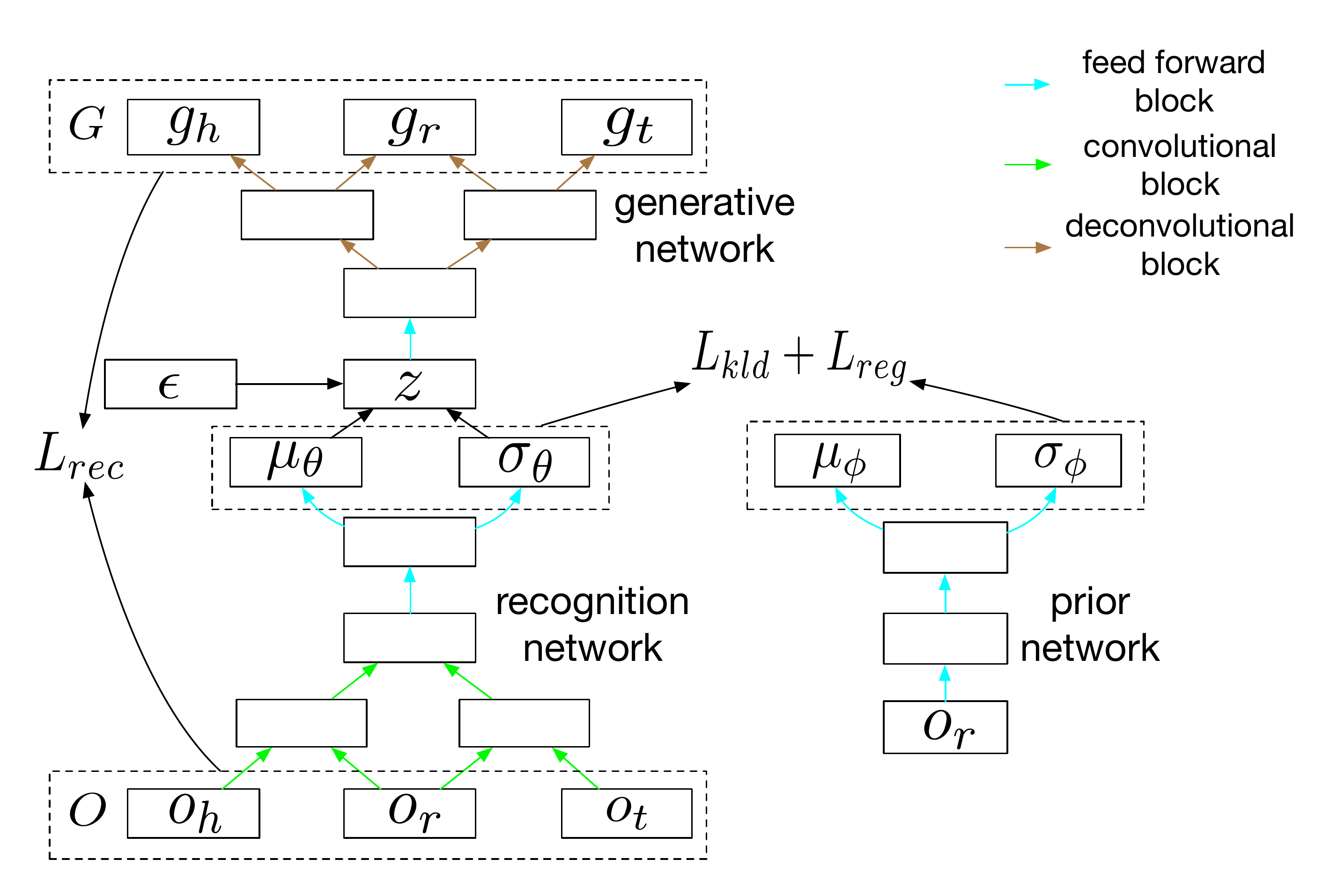}
	\caption{Structure of TCVAE. This figure should be viewed in color.}
	\label{fig:figure3}
\end{figure}

When handling KGC, learning good representation for infrequent relations and uncommon entities is difficult due to data sparsity.
However, recently some research has focused on relieving the data sparsity in few-shot learning by generating extra data with a generative model \cite{NIPS2018_7549, Wang2018LowShotLF}.
Inspired by these works, we propose a deep generative model that aims at triplet augmentation for $k$-shot learning.
Although the generative adversarial network \cite{Goodfellow:2014:GAN:2969033.2969125} is a popular model that can generate high-quality samples \cite{FridAdar2018SyntheticDA}, it suffers from an unstable learning process in our framework because of the difficult nature of Nash equilibrium and the influence of meta-learner.
On the other hand, VAE is often applied to generate samples and extract latent semantics \cite{PuGHYLSC16,LiWLRB17} due to its smooth learning procedure.
To cope with this issue, we design our triplet generator on the basis of CVAE \cite{SohnLY15} and we name it triplet CVAE (TCVAE) in this paper.
Figure 3 depicts the overall structure of TCVAE that is composed of three important probability distributions:
\begin{itemize}
	\item Variational posterior distribution $q_{\theta}(z | O)$ parameterized by $\theta$ of the recognition network.
	\item Conditional prior distribution $p_{\phi}(z | o_r)$ parameterized by $\phi$ of the prior network.
	\item Likelihood distribution $p_{\psi}(G | z, o_r)$ parameterized by $\psi$ of the generative network.
\end{itemize}

In the recognition network, there exist two layer of convolutional blocks.
Each convolutional block takes two $u$-dimensional inputs and concatenates them to form a matrix with shape $(2, u)$, so that 1-d convolution with filter width 2 can be applied to it.
Instead of directly concatenate $O = (o_h, o_r, o_t)$ to form a matrix with shape $(3, u)$ and adopt only one layer of convolutional block, such a tree structure of two consecutive layers can better capture the pairwise semantics between any two embeddings in $O$.
Likewise, two layers of the deconvolutional blocks that takes a $u$-dimensional vector as input and outputs a matrix with shape $(2, u)$ are placed in the generative network.
Besides, all feed forward blocks in TCVAE consist of an affine transformation and a non-linear mapping.

During the meta-training phase, the recognition network takes $O$ as input and learns the variational parameters $\mu_{\theta}$ and $\sigma_{\theta}$ of the variational posterior $q_{\theta}(z | O)$, where the latent semantics $z$ is assumed to follow a Gaussian distribution.
Besides, the prior network conditioning on $o_r$ computes the parameters $\mu_{\phi}$ and $\sigma_{\phi}$ of the prior distribution $p_{\phi}(z | o_r)$.
After that, the generative network samples three $u$-dimensional embeddings $G = (g_h, g_r, g_t)$ from the likelihood distribution $p_{\psi}(G | z, o_r)$.
In the generative network, firstly, the latent variable $z$ is transformed into a $u$-dimensional hidden state with the feed forward block after $z$.
Next, the first deconvolutional block receives the hidden state and outputs a matrix with shape $(2, u)$.
Finally, the second deconvolutional block receives the matrix before and outputs a matrix with shape $(3, u)$ which is denoted to $G$.
$G$ consists of three u-dimensional embeddings $(g_h, g_r, g_t)$ corresponding to three elements $(h, r, t)$ in the triplet.

Given the procedure, we are able to write down the loss terms of TCVAE: $L_{rec}$, $L_{kld}$ and $L_{rec}$.
More formally, $L_{rec}$ is the expected log-likelihood that is also the reconstruction loss between the input $O$ and the output $G$
\begin{equation}
L_{rec} = E_{q_{\theta}(z | O)} \log p_{\psi}(G | z, o_r).
\end{equation}
$L_{kld}$ is the KL-divergence between variational posterior distribution and conditional prior distribution
\begin{equation}
L_{kld} = KL(q_{\theta}(z | O) \Vert p_{\phi}(z | o_r)).
\end{equation}
And $L_{rec}$ is the regularization term for the prior network proposed in \cite{ivanov2018variational} 
\begin{equation}
L_{reg} = -\frac{\mu_{\phi}^2}{2 \sigma_{\mu}^2} + \sigma_{\sigma}(\log \sigma_{\psi} - \sigma_{\psi}),
\end{equation}
where $\sigma_{\mu} = 10000 $ and $\sigma_{\sigma} = 0.0001$ are two hyper-parameters.
There terms are jointly optimized with the loss function of KGC that we would demonstrate in the following subsection.

During the training stage of meta-testing phase, TCVAE uses prior network to compute $\mu_{\phi}$ and $\sigma_{\phi}$ given only the relation embedding $o_r^{'}$ within $O$ where $r^{'} \in R_{test}$.
Then it obtains $K$ latent variables $z$ with the following transformation
\begin{equation}
z = \mu_{\phi} + \sigma_{\phi} \odot \epsilon \quad \epsilon \sim Gaussian(0, 1).
\end{equation}
After that, $K$ embeddings of triplet $G$ can be generated from the likelihood distribution $p_{\psi}(G | z, o_r^{'})$, and $G$ is merged with $O$ to form the final embeddings $E$.
Please note that $E$ is subsequently used to compute the score $C$ with the score function $F$ in Equation \ref{eq:1}.

\subsection{Loss Function and Meta-Learner}
Following previous works, we adopt a simple strategy to compute the loss function of KGC $L_{KGC}$ in both meta-training phase and the training stage of meta-testing phase.
Given a randomly sampled relation $r$, first a positive triplet $(h, r, t)$ is sampled from all triplets of $r$.
Next, a negative triplet $(h, r, t^{'})$ can be produced by replacing $t$ with another entity $t^{'}$ in KG, where the replacement is based on a uniform negative sampling.
Note that if the negative triplet exists in KG, the negative sampling needs to be performed again.
With a positive triplet and a negative triplet, embeddings of the positive triplet $E^+$ and embeddings of the negative triplet $E^-$ can be obtained, and then two scores $C^+$ and $C^-$ can be computed respectively.
Finally, a hinge loss related to both scores is minimized for performing KGC
\begin{equation}
L_{KGC} = max(0, \gamma + C^+ - C^-),
\end{equation}
where $\gamma$ is a margin hyper-parameter greater than $0$.
Moreover, during meta-training phase, $L_{KGC}$ is also jointly optimized with TCVAE so that the overall loss $L$ is
\begin{equation}
L = L_{KGC} - L_{rec} - \lambda_1 L_{kld} - \lambda_2 L_{reg},
\end{equation}
where the negative loss terms of TCVAE are minimized, and $\lambda_1$, $\lambda_2$ are two hyper-parameters for weighting terms in the overall loss.

In order to ensure that both description encoder and TCVAE have a good generalization ability when handling infrequent relations and uncommon entities, a meta-learner is further used to optimize the overall loss $L$.
Among different directions of meta-learning, we construct the meta-learner based on Reptile since we find it has the best performance in our task.
In the context of KGC, learning with Reptile is different from previous KGC works.
During the meta-training phase, Reptile searches for an initial point $W$ within the parameter space of our framework, but such a framework may not perform well when directly used for performing KGC in $R_{test}$.
Instead, during training stage of the meta-testing phase, the framework parameters can be quickly adapted to a new point $W^{'}$ that is suitable for performing KGC given a new relation $r^{'} \in R_{test}$, and such an adaptation only needs a few training triplets of $r^{'}$ available. 
The procedure of learning with the meta-learner is depicted in Algorithm \ref{algo:meta}, where Adam \cite{Kingma2015AdamAM} is used during the $S$ inner-training steps.

\begin{algorithm}[h]
	\caption{Learning Procedure of Meta-Learner}
	\label{algo:meta}
	\begin{algorithmic}[1]
	%\State \textbf{Input:} $R_{train}$, initial model parameters $W$
	\For {iteration $= 1, 2, \dots$}
		\State Save parameters of our framework $W$
		\For {$i = 1, \dots, B$}
			\State Sample a relation $r \in R_{train}$, then sample a positive and a negative triplet of $r$
			\State Train our framework for $S$ steps using Adam with learning rate $\alpha_1$
			\State Save current parameters of our framework $W_i$ and reset them to $W$
		\EndFor
		\State \textbf{end for}
		\State Update $W \leftarrow W + \alpha_2 (\frac{1}{B} \sum_{i=1}^B W_i - W)$
	\EndFor	
	\State \textbf{end for}
	\end{algorithmic}
\end{algorithm}	

\iffalse
\begin{algorithm}[h]
	\caption{Meta-Testing Procedure}
	\begin{algorithmic}[1]
		\State \textbf{Input:} $R_{test}$, trained model parameters $W$
		\For {$r^{'} \in R_{test}$}
			\State Sample $k$ out of $C_{r^{'}}$ positive triplets
			\State Compute $O$ 
			\State Generate $K$ embeddings of triplets $G$
			\State Train model with $k + K$ data for $S$ steps using Adam and obtain new model parameters $W^{'}$			
			\State Evaluate model with remaining $C_{r^{'}} - k$ triplets
			\State Reset model parameters to $W$
		\EndFor
		\State \textbf{end for}
	\end{algorithmic}
\end{algorithm}	
\fi

\section{Experiments}
\subsection{Datasets}

\begin{table}[h]
	\centering
	\begin{tabular}{ccc}
		\hline
		 & WDtext & DBPtext \\
		\hline
		$\#$Entity & 60304 & 51768 \\
		$\#$Relation & 178 & 319 \\
		$\#$Word & 131796 & 170844 \\
		$\#R_{train}$ & 130 & 220 \\
		$\#R_{val}$ & 16 & 30 \\
		$\#R_{test}$ & 32 & 69 \\
		\makecell{Avg $\#$words} & 5.3 & 170.8 \\
		\hline
	\end{tabular}
	\caption{Statistics of datasets, where "Avg $\#$words" means the average number of words in descriptions.}
	\label{table:statistics}
\end{table}

\begin{table*}[h]
	\centering
	\begin{tabular}{lcccc|cccc}
		\hline 
		& \multicolumn{4}{c}{WDtext} & \multicolumn{4}{c}{DBPtext} \\
		\cline{2-9}
		Model & Hits@10 & Hits@5 & Hits@1 & MRR & Hits@10 & Hits@5 & Hits@1 & MRR\\
		\hline
		DKRL & 0.180 & 0.143 & 0.104 & 0.137 & 0.100 & 0.041 & 0.010 & 0.054 \\ % 2500, 500
		ConMask & \textbf{0.279} & \textbf{0.207} & 0.085 & 0.156 & 0.304 & 0.213 & 0.059 & 0.147 \\ % 2000, 1500
		GMatching & 0.095 & 0.092 & 0.090 & 0.093 & 0.194 & 0.141 & 0.100 & 0.138 \\ % 1000, 2000
		Ours-trait & 0.179 & 0.168 & 0.081 & 0.126 & 0.182 & 0.124 & 0.060 & 0.107 \\ % 200, 200
		Ours-TCVAE & 0.178 & 0.155 & 0.138 & 0.149 & 0.343 & 0.258 & 0.110 & 0.187 \\ % 400, 100
		Ours & 0.198 & 0.189 & \textbf{0.148} & \textbf{0.168} & \textbf{0.376} & \textbf{0.321} & \textbf{0.224} & \textbf{0.281} \\
		\hline
	\end{tabular} 
	\caption{One-shot KGC results on WDtext and DBPedia, where \textbf{bold} numbers indicate best results over different models on the same metric.}
	\label{table:one-shot}
\end{table*}

\begin{table*}[!]
	\centering
	\begin{tabular}{lcccc|cccc}
		\hline 
		& \multicolumn{4}{c}{WDtext} & \multicolumn{4}{c}{DBPtext} \\
		\cline{2-9}
		Model & Hits@10 & Hits@5 & Hits@1 & MRR & Hits@10 & Hits@5 & Hits@1 & MRR\\
		\hline
		DKRL & 0.202 & 0.173 & 0.122 & 0.151 & 0.099 & 0.053 & 0.012 & 0.052 \\	% 500, 1000
		ConMask & \textbf{0.303} & 0.211 & 0.107 & 0.173 & 0.333 & 0.225 & 0.077 & 0.163 \\	% 5000, 4500 
		GMatching & 0.089 & 0.088 & 0.085 & 0.088 & 0.185 & 0.146 & 0.097 & 0.138 \\ %
		Ours-trait & 0.277 & 0.221 & 0.113 & 0.170 & 0.252 & 0.174 & 0.088 & 0.146 \\ % 500, 400
		Ours-TCVAE & 0.214 & 0.198 & 0.159 & 0.184 & 0.303 & 0.227 & 0.110 & 0.174 \\ % 200, 300	
		Ours & 0.258 & \textbf{0.227} & \textbf{0.180} & \textbf{0.210} & \textbf{0.409} & \textbf{0.319} & \textbf{0.186} & \textbf{0.255} \\ %								
		\hline
	\end{tabular} 
	\caption{Four-shot KGC results on WDtext and DBPedia, where \textbf{bold} numbers indicate best results over different models on the same metric.}
	\label{table:few-shot}
\end{table*}

Existing datasets for KGC usually select triplets consisting of frequently appearing relations and common entities.
Recently two datasets focused on infrequent relations are proposed in \cite{XiongYCGW18}, but they do not contain textual descriptions for relations and entities.
To obtain datasets that fulfill the practical few-shot learning situations as investigated in the problem setting, we manually harvest triplets and their textual descriptions from Wikidata \cite{Vrandecic:2014:WFC:2661061.2629489} and DBpedia \cite{lehmann2015dbpedia}, and then we construct two datasets called WDtext and DBPtext respectively.
The statistics of the two datasets are shown in Table \ref*{table:statistics}.
Specifically, the average numbers of words in the textual descriptions have a large variation between two datasets.
In WDtext, descriptions are usually short phrases with only several words, on the other hand, descriptions in DBPtext consists of thousands of words, and we use first 200 words so that our model can be processed on GPUs.
Such a variation of description length can better reveal the performance of our model in different situations.
Besides, in order to collect enough triplets for evaluation, we select those relations whose numbers of triplets are greater than 5 and less than 1000, where the contained entities may exist only several times or even unseen during the meta-training phase.
In this way, the problem setting of infrequent relations and uncommon entities is fulfilled in our datasets.

\subsection{Experiment Setting}
We compare our model with previous KGC models that can make use of textual descriptions, namely, DKRL \cite{XieLJLS16} and ConMask \cite{ShiW18}.
We adapt their codes with our implementation.
Following the experimental protocol of \cite{ShiW18}, we also remove structural features of DKRL so that it can tackle unseen entities.
To facilitate fair comparisons, even though these models are designed without using meta-learning, for each relation in $R_{val}$ and $R_{test}$, we also sample $k$ triplets and put them into $R_{train}$ to ensure that all models make use of similar training data. 
Besides, the few-shot KGC model GMatching \cite{XiongYCGW18} is also used for comparison.
We only enable its "neighbor encoder" on WDtext because we cannot collect neighbor information from DBPedia.
Moreover, we also design two additional baselines for ablation study.
These baselines are constructed by removing specific components and keeping remaining parts in our framework.
Specifically, trait is removed in the baseline "Ours - trait" and TCVAE is removed in the baseline "Ours - TCVAE" respectively.
The experimental setting of hyper-parameters and initialization of our framework and baselines can be found in Appendix A.

To make a fair comparison, we use two categories of common metrics: mean reciprocal rank (MRR) and hits@P which is the percentage of correct tail entities ranked in the top P.
Besides, experiments are conducted using four different random seeds and we report the average results of four trials.
For each model, we select the epoch that has the best performance when evaluating on $R_{val}$ and report the corresponding results on $R_{test}$.

\begin{table}[!]
	\centering
	\begin{tabular}{cc|cc}
		\hline 
		\multicolumn{2}{c}{WDtext} & \multicolumn{2}{c}{DBPtext} \\
		\hline
		$\#$Gen & MRR & $\#$Gen & MRR \\
		\hline
		0 & 0.149 & 0 & 0.187 \\ 
		2 & 0.166 & 16 & 0.247 \\ 
		4 & 0.172 & 32 & 0.274 \\ 
		8 & 0.179 & 64 & 0.304 \\ 
		16 & 0.146  & 128 & 0.330 \\ 
		32 & 0.144 & 256 & 0.256 \\ 
		\hline
	\end{tabular} 
	\caption{MRR results of our framework when using different number of augmentation in TCVAE, where $\#$Gen means the number of triplets being generated.}
	\label{table:augmentation}
\end{table}

\subsection{Results of One-Shot Learning}
Firstly, we compare our overall framework with baselines by conducting an one-shot KGC experiment, which is the most difficult case in few-shot learning.
The results of our overall framework and baselines are shown in Table \ref{table:one-shot}, where we find that our overall framework outperforms the baselines on most metrics.
For WDtext, ConMask is a strong baseline and has a better result on Hits@10 and Hits@5, but it performs worse on Hits@1 and MRR compared to our framework.
On the other hand, our overall framework outperforms for a large margin compared to all baselines on DBPtext.
Since the main difference between the two datasets are the average length of descriptions, we can observe that our framework has a better performance when dealing with long textual descriptions.
Besides, the results of the two ablation baselines are significantly worse than our overall framework, and thus we can see both components play an important role in our framework.

\subsection{Results of Few-Shot Learning}
In real-world KGs, few-shot learning of infrequent relations and uncommon entities are more common than one-shot scenario, so we also conduct four-shot KGC as another experiment.
We use the same baselines and metrics as that in the one-shot KGC experiment.
The results are shown in Table \ref{table:few-shot}, where we can see that our overall framework also has the best performance on different comparisons except for Hits@10 on WDtext.
By comparing our overall framework with two ablation baselines, the importance of traits and triplet augmentation is demonstrated again.
Furthermore, when compared with the previous one-shot KGC results, all baselines in this experiment are able to make use of the extra training data to improve their performances.
Our framework also performs better on all metrics in WDtext, but it performs inferior on some metrics in DBPtext when compared with the one-shot scenario.
One possible reason is descriptions in DBPtext is longer and more complex than that in WDtext, and thus four descriptions of training triplets are too diverse to learn a good representation.

\subsection{Analysis of Triplet Generation}
As shown in the previous KGC experiments, the performance of our framework heavily depends on the triplet generation provided by TCVAE.
In this subsection, we further explore the effect of triplet generation by comparing the MRR result of our overall framework with different number of generated triplets in one-shot scenario, where the triplet augmentation is particularly helpful.
For this analysis, we only conduct experiment and report the result with one trial for simplicity.
The results are shown in Table \ref{table:augmentation}, and we can conclude that a proper data augmentation does enhance the performance of our framework when training data available is scarce.
Besides, the appropriate number of triplets being generated varies from one dataset to another, and too many or too few generation leads to an inferior performance.
In WDtext, generating 8 extra triplets enhances the performance most, but generating 128 triplets is better in DBPtext.
One reason for such a difference is that DBPtext has longer descriptions which also increase the variance of the generated triplets, and thus generating more triplets is necessary to learn a stable representation.

\section{Conclusions}
We consider a new type of KGC where infrequent relations and uncommon entities need to be jointly handled, and we formulate it as a few-shot KGC problem.
To tackle the problem, we propose a novel concept "trait" and adopt it to extract relation-specific information from entity descriptions.
Besides, we also design a triplet generator and a meta-learning framework based on Reptile to deal with the issue of few-shot KGC.
Moreover, we also conduct two new datasets that focus on this problem setting.
The experiments of both one-shot and four-shot scenarios show that our framework has a better performance compared to other baselines.

%\section*{Acknowledgments}
%\import{./sections/}{8_acknowledgements.tex}

\clearpage
\bibliography{emnlp-ijcnlp-2019}
\bibliographystyle{acl_natbib}
%\fi

%\iffalse
\clearpage
\appendix
\section{Hyper-parameter Setting}
\label{appendix:setting}

For our framework, word embeddings and other trainable parameters are randomly initialized with Xavier \cite{Glorot2010UnderstandingTD}.
The dimension of word embedding, the dimension of hidden layers in both CNN and TCVAE and the number of convolutional filters in both CNN and TCVAE are 100.
The non-linear mapping function is $\tanh$ everywhere.
In CNN, the width of convolutional filters is 3 and the number of memory is 128.
Moreover, the dimension of pooling stride is 2 for WDtext and 4 for DBPtext.
In TCVAE, the dimension of latent layer is 50.
Other hyper-parameters can be viewed in Table \ref{table:hyper-parameter}.
In order to make a fair comparison, for DKRL and ConMask, their dimensions of word embedding and hidden layer, parameters of convolutional filters and strides of pooling layer are same as what we used in our framework, and other hyper-parameters are set to ones mentioned in their papers.
For GMatching, we randomly initialize its embeddings like other models, and other hyper-parameters are their default values.

\begin{table*}[bp]
	\centering
	\begin{tabular}{cccc}
		\hline
		Name of hyper-parameter & Symbol in paper & Value in WDtext & Value in DBPtext \\
		\hline
		number of generated triplet& K & 8 & 128 \\
		batch size & B & \multicolumn{2}{c}{8} \\
		number of layers in CNN& N & \multicolumn{2}{c}{3} \\
		number of steps in inner iteration& S & \multicolumn{2}{c}{5} \\
		weight of $L_{kld}$ & $\lambda_1$ & \multicolumn{2}{c}{1.0} \\
		weight of $L_{reg}$ & $\lambda_2$ & \multicolumn{2}{c}{1.0} \\
		learning rate of inner iteration & $\alpha_1$ & \multicolumn{2}{c}{0.001} \\
		learning rate of Reptile &$ \alpha_2$ & \multicolumn{2}{c}{0.001} \\
		\iffalse
		dimension of word embedding & & \multicolumn{2}{c}{100} \\
		dimension of hidden layers in CNN & & \multicolumn{2}{c}{100} \\
		number of convolutional filters in CNN & &\multicolumn{2}{c}{100} \\
		dimension of convolutional filters in CNN & & \multicolumn{2}{c}{3} \\
		dimension of pooling stride in CNN & & 2 & 4 \\
		number of memory in CNN & & \multicolumn{2}{c}{128} \\
		number of convolutional filters in TCVAE & &\multicolumn{2}{c}{100} \\
		dimension of hidden layers in TCVAE & & \multicolumn{2}{c}{100} \\
		dimension of latent layer in TCVAE & & \multicolumn{2}{c}{50} \\		
		non-linear mapping function & & \multicolumn{2}{c}{$\tanh$} \\
		\fi
		\hline
	\end{tabular}
	\caption{Hyper-parameters of our framework.}
	\label{table:hyper-parameter}
\end{table*}

%\clearpage
%\bibliography{emnlp-ijcnlp-2019}
%\bibliographystyle{acl_natbib}
%\fi
\end{document}